\documentclass[10pt,twocolumn,letterpaper]{article}

\usepackage{cvpr}
\usepackage{times}
\usepackage{epsfig}
\usepackage{graphicx}
\usepackage{amsmath}
\usepackage{amssymb}

\usepackage{url}
\usepackage{booktabs}
\usepackage{nidanfloat}
\usepackage{multirow}
\usepackage{slashbox}

\usepackage[pagebackref=true,breaklinks=true,letterpaper=true,colorlinks,bookmarks=false]{hyperref}

\cvprfinalcopy 

\def\cvprPaperID{815} 

\ifcvprfinal\fi
\begin{document}

\title{Recognizing Activities of Daily Living with a Wrist-mounted Camera}

\author{Katsunori Ohnishi, Atsushi Kanehira, Asako Kanezaki, Tatsuya Harada\\
Graduate School of Information Science and Technology, The University of Tokyo\\
{\tt\small \{ohnishi, kanehira, kanezaki, harada\}@mi.t.u-tokyo.ac.jp}}

\maketitle
\thispagestyle{empty}

\begin{abstract}
We present a novel dataset and a novel algorithm for recognizing activities of daily living (ADL) from a first-person wearable camera. Handled objects are crucially important for egocentric ADL recognition. For specific examination of objects related to users' actions separately from other objects in an environment, many previous works have addressed the detection of handled objects in images captured from head-mounted and chest-mounted cameras. Nevertheless, detecting handled objects is not always easy because they tend to appear small in images. They can be occluded by a user's body. As described herein, we mount a camera on a user's wrist. A wrist-mounted camera can capture handled objects at a large scale, and thus it enables us to skip the object detection process. To compare a wrist-mounted camera and a head-mounted camera, we also developed a novel and publicly available dataset \footnote{http://www.mi.t.u-tokyo.ac.jp/static/projects/miladl/} that includes videos and annotations of daily activities captured simultaneously by both cameras. Additionally, we propose a discriminative video representation that retains spatial and temporal information after encoding the frame descriptors extracted by convolutional neural networks (CNN). 

\end{abstract}

\def\vector#1{\mbox{\boldmath $#1$}}

\section{Introduction}
Recently, activity recognition from first-person camera views has been attracting increasing interest, motivated by advances in wearable device technology. Recognition of activities of daily living (ADL) from first-person views is an important task related to activity recognition. ADL are basic activities in a typical human life such as ``making coffee'' or ``cutting paper.'' If the system recognizes ADL properly, then it is applicable to nursing services, rehabilitation, and lifestyle habit improvements.

To recognize ADL, it is important to examine objects undergoing hand manipulation specifically. For example, a cleaning activity might be recognized only by recognizing that a user is using a vacuum cleaner. One can also recognize coffee-making activity if it is observed that a user is handling a mug and coffee beans. Pirsiavash and Ramanan \cite{pirsiavash2012detecting} described the importance of recognizing handled objects for ADL recognition. They developed an ADL dataset collected using a chest-mounted camera. Then, they implemented ADL recognition in different homes by detecting the user's hands and handled objects. The result suggests the crucial importance of detecting the handled objects properly in various environments for ADL recognition.

\begin{figure}[t]
\begin{center}
   \includegraphics[width=\linewidth]{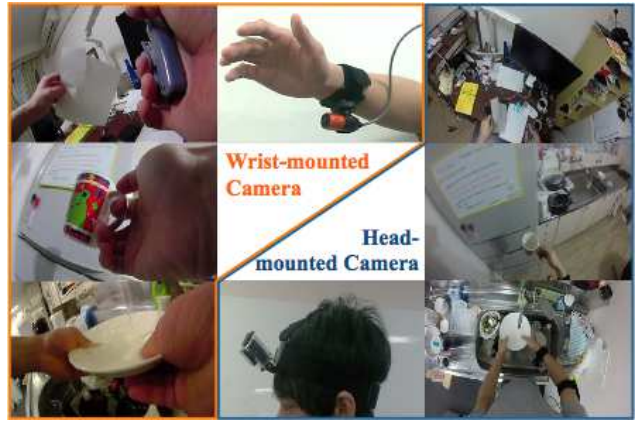}
\end{center}
\label{fig:represents}
\vspace{-3mm}
   \caption{Activities of daily living (ADL) captured by a wrist-mounted camera ({\bf left}) and a head-mounted camera ({\bf right}).}
\vspace{-2mm}
\end{figure}

In a view from a head-mounted camera or a chest-mounted camera, handled objects are captured at a small scale in various positions. Furthermore, many non-handled objects also appear in the captured image. Consequently, many studies have examined hand detection or gaze prediction to develop a means of discerning picked-up and handled objects from other detected objects. However, such approaches entail the following difficulties: (a) Despite the advances in object-detection techniques, object detection is not an easy task in various environments. (b) Discerning a handled object from detected objects with hand detection or gaze prediction is not an easy task. (c) To train object detectors, it is necessary to collect numerous images with bounding boxes. Building a high-quality dataset with bounding boxes requires a considerable amount of labor, which hinders us from expanding a dataset.

To train an ADL recognition system without pixel-level annotations or bounding boxes of objects, we consider mounting a wearable camera on the wrist of the user's dominant arm because the objects are handled mainly by the user's dominant hand. We designate this camera as a wrist-mounted camera in this paper. For ADL recognition, a wrist-mounted camera has numerous advantages over a head-mounted camera or a chest-mounted camera: (a) Wrist-mounted cameras can capture a large image of the handled objects. (b) Because handled objects are close to a user's dominant hand, the object positions are restricted in the images from the wrist-mounted camera. (c) Because of the above reasons, we can skip object detection and do not need a dataset of ADL with bounding boxes. We only need a dataset of ADL with annotations about the activity time segments in the videos.

We also propose a recognition system for videos captured by a wrist-mounted camera that has strong spatial bias and weak temporal bias. As shown in Figure \ref{fig:wrist_mean}, an image captured by a wrist-mounted camera has strong spatial bias, meaning that hand-manipulated objects tend to be located at the central area. In addition, the order of manipulated objects is mostly fixed for each action, which we call ``weak temporal bias.'' The state-of-the-art video representation, which extracts local features containing spatial information from pre-trained convolutional neural networks (CNNs), strongly loses spatial information and completely loses temporal information after encoding. Therefore, we also propose a novel video representation that retains spatial and temporal information after encoding to consider the above mentioned biases.

Our three contributions are the following:
\begin{enumerate}
  \setlength{\parskip}{-1mm} 
  \setlength{\itemsep}{1mm} 
\item We propose the use of a wrist-mounted camera for ADL recognition instead of a head-mounted camera or a chest-mounted camera. 
\item We propose a discriminative video representation that retains spatial and temporal information. This is a method for the dataset captured from a wrist-mounted camera that has a large bias of spatial information and a small bias of temporal information.
\item We developed a novel and publicly available dataset that includes videos and annotations of ADL captured from a head-mounted camera and a wrist-mounted camera simultaneously. 
\end{enumerate}

\vspace{-6mm}
\section{Related work}
\vspace{-1mm}
\subsection{Egocentric vision for ADL recognition}
\vspace{-1mm}
Various approaches for ADL recognition based on handled objects have been proposed \cite{patterson2005fine, stikic2008adl, wu2007scalable}. Because wearable devices with cameras such as GoPro and Google Glasses have been developed recently, ADL recognition with viewpoint cameras has received a considerable amount of attention. Some works on egocentric ADL recognition have achieved results in {\it a single environment}, such as a kitchen or an office \cite{hanheide2006action,fathi2011learning,li2013learning,li2015delving}.

For more practical settings, Pirsiavash and Ramanan \cite{pirsiavash2012detecting} estimated the type of a handled object by detecting the object and arm from a wearable camera's viewpoint. They demonstrated that action recognition performs well in {\it diverse environments}. However, it is necessary to provide positional information of all objects in all frames of the video at the time of learning. In addition, detecting an entire handled object itself is still difficult. Although their dataset has various annotations, such as type of activity and duration of its completion, as well as type of an object and its location, it took over 1 month to label various annotations by 10 part-time annotators. Consequently, expanding the dataset is not practical. In a more practical setting, an ADL recognition system that uses wearable devices in diverse environments should be trained with labels obtained by simpler annotation methods.

\subsection{Video representation for action recognition}
\vspace{-2mm}
Video representation has been well studied in the action recognition domain. Some deep-learning approaches for action recognition have been proposed \cite{ng2015beyond, tran2014c3d}. However, these approaches require the use of large-scale video datasets (e.g. Sports 1M \cite{karpathy2014large}), which are difficult to address and which require enormous amounts of time for the whole learning process.

{\bfseries Motion features:} The general pipeline to obtain a video representation for action recognition models the distribution of local features from training videos. Local features representing motion information (e.g., HOG \cite{dalal2005histograms}, HOF \cite{lucas1981iterative}, and MBH \cite{dalal2006human}) are usually used. The combination of local features and improved dense trajectory (iDT) \cite{wang2013action}, which compensates for camera motion, is the de facto standard. It has shown great performance for action recognition \cite{wang2013dense}.

{\bfseries CNN descriptors:} CNN has achieved superior results to the standard pipeline for object recognition \cite{krizhevsky2012imagenet}. Jain \etal \cite{jainuniversity} brought CNN to action recognition. They obtain the state of a fully connected layer from each frame in videos and calculate the video representation by averaging all CNN features. Their method therefore exhibits performance that is surprisingly comparable to the combination of iDT, MBH, and Fisher vector (FV) \cite{perronnin2007fisher}. To obtain more discriminative features containing spatial information, Xu \etal \cite{xu2015discriminative} proposed the extraction of latent concept descriptors (LCDs) from the pool$_5$ layer and the application of VLAD \cite{jegou2010aggregating} instead of averaging. However, spatial information is ignored when applying VLAD. Since our task is an intermediate task of action recognition and object recognition, we developed the CNN-based video representation above to design the video representation for ADL recognition from wrist-mounted cameras, which have strong spatial information bias.

\section{Wrist-mounted cameras}
\vspace{-3mm}
\label{sec:wmc}
Some works in the area of interface research have shed light on wrist-mounted cameras \cite{vardy1999wristcam, kim2012digits}. In ADL recognition, Maekawa \etal \cite{maekawa2010object} conducted multimodal ADL recognition using a wrist-mounted device that has a camera, microphone, acceleration sensor, illuminance meter, and digital compass. However, the color histogram alone is used as an image feature. This system is too simple to identify handled objects. Wrist-mounted cameras have never been evaluated carefully in ADL recognition. Therefore, we discuss the superiority of wrist-mounted cameras in this section.

\begin{figure}[t]
\begin{center}
\includegraphics[width=0.85\linewidth]{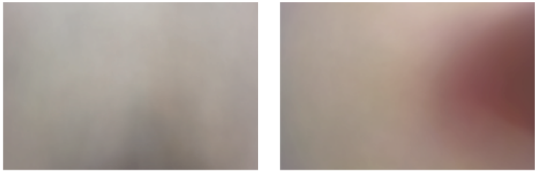}
\end{center}
\vspace{-2mm}
   \caption{Mean images of a head-mounted camera ({\bf left}) and a wrist-mounted camera ({\bf right}). Skin pixels are visible on the right side of the wrist-mounted camera image, although we cannot see anything in the head-mounted camera image. This implies that the user's hand always appears in the right side and handled objects appear near the center of the wrist-mounted camera image.}
\label{fig:wrist_mean}
\vspace{-2mm}
\end{figure}

Wrist-mounted cameras capture handled objects very closely, as shown in Figure \ref{fig:represents}. In addition, as shown in Figure \ref{fig:wrist_mean}, the user's hand invariably appears on the right side of the image captured by a wrist-mounted camera, unlike that by a head-mounted camera. This trend of wrist-mounted cameras also means that handled objects always appear near the center of the captured image. 
Because of these strong spatial biases, we can recognize handled objects well even without manually annotating the bounding boxes of objects in the dataset. We need only to annotate the time segments of the activities. Wrist-mounted cameras have limitations: they cannot take pictures of human faces or recognize posture-defined actions such as ``jumping'' or ``skipping.'' Although there are such limitations, wrist-mounted cameras are more suitable for recognizing ADL, which mostly involves object manipulation.

As another feature, wrist-mounted cameras can process large motions. Because the handled objects and a wrist-mounted camera move together, other irrelevant parts, which move relative to the wrist-mounted camera, are blurred, whereas the handled objects are captured clearly. In addition, the objects, while moving, appear as static objects in the camera view, which enables robust recognition. This blurring effect is better obtained by setting the focal length to 10--30 cm.

\vspace{-1mm}
\section{Video representation}
\vspace{-2mm}
We propose a new video representation based on LCDs \cite{xu2015discriminative} to take advantage of the strong spatial bias and weak temporal bias of the video captured by a wrist-mounted camera. Although LCD retains the spatial information in each frame at the descriptor level, spatial information and temporal information are dropped when the descriptors are encoded and aggregated into a video representation. However, the video captured by a wrist-mounted camera has a strong spatial bias, as described in Section \ref{sec:wmc}. Although not as strong as spatial bias, temporal bias also exists because the order of handling objects is fixed roughly in each action class. Therefore, we use the benefits of these spatial and temporal biases specifically for a wrist-mounted camera and ADL.

Our method encodes LCDs at each location in all frames into single VLAD \cite{jegou2010aggregating} vectors and optimizes the weights for the VLAD vectors to aggregate them into a video representation. The weight for a VLAD vector extracted from each location is designated as a spatial weight. Furthermore, we propose a method that divides a video into short sequences and optimizes the temporal weights for aggregating descriptors. Here, we describe the original LCD in Section \ref{sec:LCD}, the proposed method to optimize spatial weights in Section \ref{sec:DSAR}, and the proposed method to optimize spatial and temporal weights in Section \ref{sec:DSTAR}.

\vspace{-1mm}
\subsection{CNN latent concept descriptors}
\label{sec:LCD}
\vspace{-2mm}
Latent concept descriptors \cite{xu2015discriminative} constitute a state-of-the-art video representation using CNN, which is obtained as follows. (i) Given a video $\mathcal{E}$ including $T$ frames $\mathcal{E}=\{I_1, I_2, \cdots, I_T\}$, each frame is input to VGG net \cite{Simonyan14c} pre-trained on the ImageNet2012 dataset \cite{deng2009imagenet} to obtain the pool$_5$ layer's output. The dimension of pool$_5$ features is $a \times a \times M$, where $a$ is the size of the filtered images of the last pooling layer and $M$ is the number of convolutional filters in the last convolutional layer (in the case of VGG net, $a=7$ and $M=512$). (ii) The responses of $M$ filters are concatenated for the respective locations of the pool$_5$ layer.

Then, a set of $a^2$ descriptors $\vector{f}^{t}_{(i,j)} \in \mathbb{R}^M$ is obtained from the $t$-th frame as follows.

\begin{equation}
\mathcal{F}^t=\{ {\vector{f}_{(1,1)}^t },{\vector{f}_{(1,2)}^t },\ldots,{\vector{f}_{(a,a)}^t } \}.
\end{equation}
(iii) All descriptors in $\{\mathcal{F}^1,\dots,\mathcal{F}^T\}$ are encoded with VLAD into a video representation $\vector{v}$. 
Letting $\{ \vector{c}_1, \dots, \vector{c}_K \}$ denote a set of $K$ coarse centers obtained by K-means, we obtain $\vector{u}_k$ $(k=1,\dots,K)$ as follows:
\begin{equation}
  \vector{u}_k = \sum_{ (t,i,j) \in \{ (t,i,j) | {\rm NN}\left(\vector{f}^{t}_{(i,j)}\right) = \vector{c}_k  \} } ( \vector{f}^{t}_{(i,j)} - \vector{c}_k ),
\vspace{-2mm}
\end{equation}
Therein, ${\rm NN}\left(\vector{f}^{t}_{(i,j)}\right)$ represents the nearest center of $\vector{f}_{(i,j)}$.
Then, $\vector{v}$ is obtained as an $MK$-dimensional VLAD encoding vector by concatenating $\vector{u}_k$ over all $K$ centers.
(iv) Finally, $\vector{v}$ is normalized by power and L2 normalization with intra-normalization \cite{arandjelovic2013all}.

\begin{figure*}[tb]
\centering
\includegraphics[width=0.9\linewidth]{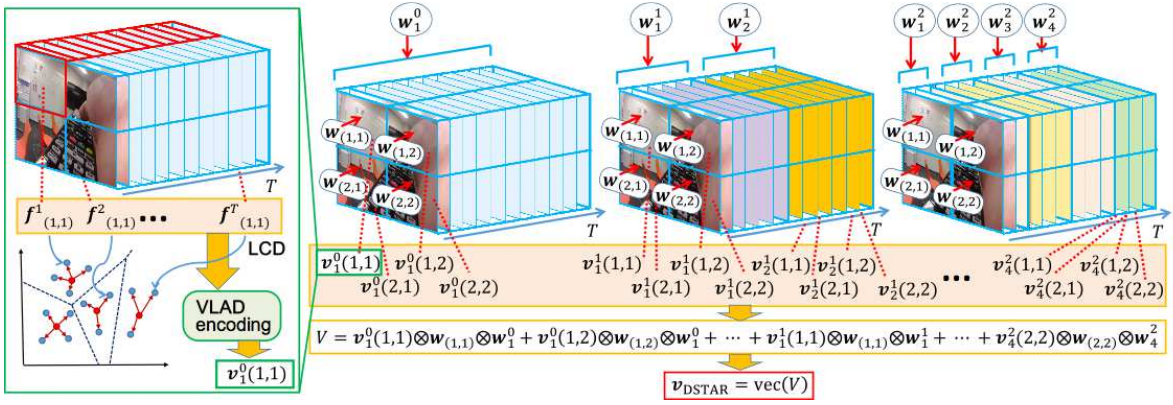}
\caption{Illustration of the proposed video representation $\vector{v}_{\rm DSTAR}$. For this example, we set $a=2$ and $L=2$.}
\label{fig:system}
\vspace{-4mm}
\end{figure*}

\vspace{-1mm}
\subsection{Discriminative spatial aggregated latent concept descriptors}
\label{sec:DSAR}
\vspace{-1mm}
The original LCD \cite{xu2015discriminative} drops the spatial information in the process of VLAD encoding because the descriptors are equally weighted when they are encoded with VLAD into a video representation $\vector{v}$. We introduce spatial weights for the VLAD encoding vectors distinguished by their locations when we aggregate them into a final video representation. Because of the spatial weights, we can address the spatial bias such that the center area is more important than its surroundings, for example, because hand-manipulated objects tend to be located at the center view of a wrist-mounted camera. Specifically, we obtain a video representation $\vector{v}(i,j)$ for each cell $(i,j)$ over all the $T$ frames by encoding the descriptors in a set $\mathcal{F}_{(i,j)}=\{ {\vector{f}_{(i,j)}^1 },\ldots,{\vector{f}_{(i,j)}^T } \}$. 
In the same manner as in \cite{xu2015discriminative}, $\vector{v}(i,j)$ is normalized by power and L2 normalization with intra-normalization.
Letting $\vector{w}_{(i,j)} \in \mathbb{R}^{N_{\rm sp}}$ denote an $N_{\rm sp}$-dimensional weight vector, we obtain a weighted sum of $\vector{v}(i,j)$ as
\begin{equation}
\vspace{-3mm}
  V = \sum_{i=1}^a \sum_{j=1}^a \vector{v}(i,j) \vector{w}_{( i, j )}^\top = V_{\rm sp} W_{\rm sp}, \label{eq:d-spr}
\end{equation}
where $V_{\rm sp} \in \mathbb{R}^{MK \times a^2}$ and $W_{\rm sp} \in \mathbb{R}^{a^2 \times N_{\rm sp}}$ are defined as shown below.
\begin{eqnarray}
  V_{\rm sp} &=& ( \vector{v}(1,1), \vector{v}(1,2), \ldots, \vector{v}(a,a)), \\
  W_{\rm sp} &=& ( \vector{w}_{(1,1)}, \vector{w}_{(1,2)}, \ldots, \vector{w}_{(a,a)})^\top.
\end{eqnarray}

As described in this paper, we obtain $W_{\rm sp}$ by arranging $N_{\rm sp}$ eigenvectors $\vector{x} \in \mathbb{R}^{a^2}$ obtained by partial least squares (PLS) in $N_{\rm sp}$ rows. Note that PLS is a method that can extract common information between sets of observed features. Therefore, $N_{sp}$ represents how many eigenvectors we use were obtained from PLS. Details related to computing the eigenvectors are given in the Supplemental Materials. Finally, we obtain a video representation $\vector{v}_{\rm DSAR} \in \mathbb{R}^{MKN_{\rm sp}}$, which is called discriminative spatial aggregated LCDs (DSAR), by concatenating all elements in $V \in \mathbb{R}^{MK \times N_{\rm sp}}$ in (\ref{eq:d-spr}).
Here, $\vector{v}_{\rm DSAR}$ is normalized by power and L2 normalization.

The idea to use the eigenvectors obtained by PLS as spatial weights was derived from the discriminative spatial pyramid representation (D-SPR) \cite{harada2011discriminative}. Consequently, the proposed method described in this section can be regarded as the combination of LCD and D-SPR. Our method optimizes the weights for $a \times a$ cells in the output of pool$_5$, whereas D-SPR optimizes the weights for areas in a spatial pyramid of each frame.

\vspace{-2mm}
\subsection{Discriminative spatiotemporal aggregated latent concept descriptors}
\label{sec:DSTAR}
\vspace{-2mm}

The wrist-mounted camera dataset not only has a strong spatial information bias, but also has a weak temporal information bias. The original LCD described in Section \ref{sec:LCD} and the proposed DSAR described in Section \ref{sec:DSAR} lose temporal information in the process of VLAD encoding. Inspired by the idea of spatiotemporal pyramids \cite{laptev2008learning}, we introduce temporal weights for the VLAD encoding vectors distinguished by their time stamps when we aggregate them into a final video representation. Because of the temporal weights, we can assign the importance of each frame into a whole video representation. Specifically, we split a video into $2^l$ sequences consisting of equal numbers of frames $T/2^l$. The $s$-th sequence is a set of frames $\{I_{(s-1)T/2^l+1},\ldots,I_{sT/2^l}\}$ $(s = 1, \ldots, 2^l)$. Then, we obtain a video representation $\vector{v}^l_s(i,j)$ for each cell $(i,j)$ over all the $T/2^l$ frames in the $s$-th sequence by encoding the descriptors in a set $\mathcal{F}_{(i,j)}^s=\{ {\vector{f}_{(i,j)}^{(s-1)T/2^l+1} },\ldots,{\vector{f}_{(i,j)}^{sT/2^l} } \}$. Here, we consider multiple levels of the splitting $(l=0,\ldots,L)$ such that we obtain a set of $\vector{v}^l_s(i,j)$ as follows:
\begin{eqnarray}
  \mathcal{V} = \{ \vector{v}^0_1(i,j),\vector{v}^1_1(i,j),\ldots,\vector{v}^L_1(i,j),\ldots,\vector{v}^L_{2^L}(i,j) \} \nonumber \\
  (i=1,\ldots,a,j=1,\ldots,a).
\end{eqnarray}
Again, $\vector{v}^l_s(i,j)$ is normalized by power and L2 normalization with intra-normalization.
Letting $\vector{w}^l_s \in \mathbb{R}^{N_{\rm tmp}}$ denote an $N_{\rm tmp}$-dimensional weight vector, we obtain a weighted sum of $\vector{v}^l_s(i,j)$ as follows:
\begin{equation}
\vspace{-3mm}
  V = \sum_{i=1}^a \sum_{j=1}^a \sum_{l=0}^L \sum_{s=1}^{2^l} \vector{v}^l_s{(i,j)} \otimes \vector{w}_{( i, j )} \otimes \vector{w}^l_s. \label{eq:proposed}
\end{equation}
Here, we define $V(i,j) \in \mathbb{R}^{MK \times d}$, $V_s^l \in \mathbb{R}^{MK \times a^2}$, and $W_{\rm tmp} \in \mathbb{R}^{d \times N_{\rm tmp}}$, where $d = \sum_{l=0}^L 2^l = 2^{L+1}-1$ as shown below.
\begin{eqnarray}
  \hspace{-4.5mm} V(i,j) \hspace{-3mm} &=& \hspace{-3mm} ( \vector{v}^0_1(i,j), \vector{v}^1_1(i,j), \ldots, \vector{v}^L_1(i,j), \ldots, \vector{v}^L_{2^L}(i,j)), \\
  \hspace{-4.5mm} V_s^l \hspace{-3mm} &=& \hspace{-3mm} ( \vector{v}_s^l(1,1), \vector{v}_s^l(1,2), \ldots, \vector{v}_s^l(a,a)), \\
  \hspace{-4.5mm} W_{\rm tmp} \hspace{-3mm} &=& \hspace{-3mm} ( \vector{w}^0_1,\vector{w}^1_1,\ldots,\vector{w}^L_1,\ldots,\vector{w}^L_{2^L} )^\top.
\end{eqnarray}

As described in this paper, we optimize spatial weights $W_{\rm sp}$ and temporal weights $W_{\rm tmp}$ iteratively and alternately.
Specifically, we repeat the following two steps.

\vspace{-4mm}
\subsubsection*{Step 1: optimizing $W_{\rm sp}$}
\vspace{-2mm}
In this step, we fix $W_{\rm tmp}$ and optimize $W_{\rm sp}$. We obtain $MKN_{\rm tmp}$-dimensional vectors $\vector{g}_{(i,j)}$ by concatenating all the elements in $V(i,j)W_{\rm tmp}$. Letting $V' \in \mathbb{R}^{MKN_{\rm tmp} \times a^2}$ denote $( \vector{g}_{(1,1)}, \vector{g}_{(1,2)}, \ldots, \vector{g}_{(a,a)} )$, 
we can rewrite (\ref{eq:proposed}) as
\begin{equation}
  V = V' W_{\rm sp}.
\end{equation}
This formulation is identical to (\ref{eq:d-spr}). Therefore, we optimize $W_{\rm sp}$ in the manner described in Section \ref{sec:DSAR}.

\vspace{-4mm}
\subsubsection*{Step 2: optimizing $W_{\rm tmp}$}
\vspace{-2mm}
In this step, we fix $W_{\rm sp}$ and optimize $W_{\rm tmp}$. We obtain $MKN_{\rm sp}$-dimensional vectors $\vector{h}^l_s$ by concatenating all the elements in $V^l_sW_{\rm sp}$. Letting $V'' \in \mathbb{R}^{MKN_{\rm sp} \times d}$ denote $( \vector{h}^0_1, \vector{h}^1_1, \ldots, \vector{h}^L_1, \ldots, \vector{h}^L_{2^L} )$, we can then
rewrite (\ref{eq:proposed}) as
\begin{equation}
  V = V'' W_{\rm tmp}.
\end{equation}
This formulation is identical to (\ref{eq:d-spr}). Therefore, we optimize $W_{\rm tmp}$ in the manner presented in Section \ref{sec:DSAR}.

We iterate Step 1 and Step 2 several times. 
Finally, we obtain a video representation $\vector{v}_{\rm DSTAR} \in \mathbb{R}^{MKN_{\rm sp}N_{\rm tmp}}$, which is called discriminative spatiotemporal aggregated LCDs (DSTAR), by concatenating all the elements in $V \in \mathbb{R}^{MK \times N_{\rm sp} \times N_{\rm tmp}}$ in (\ref{eq:proposed}) with power and L2 normalization.
An illustration of $\vector{v}_{\rm DSTAR}$ is shown in Figure \ref{fig:system}.

\vspace{-2mm}
\section{Dataset: MILADL}
\vspace{-2mm}
We created a new ADL dataset that uses both a wrist-mounted camera and a head-mounted camera because there are as yet no published ADL datasets that use wrist-mounted cameras. In this section, we present the details of our dataset.

Note that it is also important to compare a wrist-mounted camera with a chest-mounted camera instead of with a head-mounted one since a chest-mounted camera is closer to the user's hands. We encourage to compare wrist-mounting to other mountings \cite{mayol2009choice} for ADL recognition as future work.

\vspace{-2mm}
\subsection{Activity class}
\vspace{-2mm}
We chose activity classes by referring to previous studies of ADL \cite{pirsiavash2012detecting, tapia2007portable, galasko1997inventory}. First, we removed some classes that many users were reluctant to record on video such as ``brushing teeth'' and ``laundry.'' Next, to introduce more variety into our dataset, we added some actions referring to other ADL recognition studies \cite{tapia2007portable} and an evaluation of Alzheimer rehabilitation \cite{galasko1997inventory}. As Table \ref{tab:action_tables} shows, we strove to recognize 23 ADL classes in this study. Detailed information is given in the Supplemental Materials.

\begin{table}[tb]
 \begin{center}
	\small
  \begin{tabular}{|c|c|c|}
\hline 
 \multirow{2}{*}{Action name}	& Mean of		&Number of \\ 
						& length (s)	&occurrences \\ \hline
vacuuming 				&38.1		&17		\\ \hline
empty trash				&11.5		&22		\\ \hline
wipe desk					&35.7		&23		\\ \hline
turn on air-conditioner		&6.9			&27		\\ \hline
open and close door			&6.4			&34		\\ \hline
make coffee				&88.2		&24		\\ \hline
make tea					&70.0		&22		\\ \hline
wash dishes				&31.3		&29		\\ \hline
dry dishes					&15.7		&27		\\ \hline
use microwave				&33.7		&26		\\ \hline
use refrigerator				&6.8			&42		\\ \hline
wash hands				&11.9		&32		\\ \hline
dry hands					&7.9			&29		\\ \hline
drink water from a bottle		&13.8		&26		\\ \hline
drink water from a cup		&7.8			&31		\\ \hline
read book					&28.5		&28		\\ \hline
write on paper				&16.6		&29		\\ \hline
open and close drawer		&6.0			&30		\\ \hline
cut paper					&14.7		&28		\\ \hline
staple paper				&7.8			&28		\\ \hline
fold origami				&68.7		&23		\\ \hline
use smartphone			&23.1		&29		\\ \hline
watch TV					&21.6		&22		\\ \hline
  \end{tabular}
  \caption{Duration of each class and the distribution of the 23 classes in our dataset.}
 \label{tab:action_tables}
 \end{center}
 \vspace{-10mm}
\end{table}

\vspace{-1mm}
\subsection{Collection and annotation}
\begin{figure}[tbp]
\centering
\includegraphics[width=0.75\linewidth]{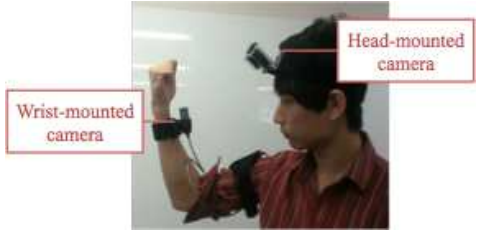}
\caption{Wearing a head-mounted camera and a wrist-mounted camera}
\label{fig:wearable}
\vspace{-3mm}
\end{figure}

\vspace{-2mm}
To assemble the dataset, we used a GoPro HERO3+ \footnote{http://jp.shop.gopro.com/cameras} as the head-mounted camera and an HX-A100 \footnote{http://panasonic.jp/wearable/a100} as the wrist-mounted camera. Each user wore these two cameras, as shown in Figure \ref{fig:wearable}. Each user therefore recorded two videos simultaneously. As in a previous ADL egocentric dataset \cite{pirsiavash2012detecting}, we did not instruct the users in detail how to act to obtain realistic data. After taking videos, all users manually annotated the duration and the action class in their own videos. The definition of an action includes some initial and final actions related to the action. For example, the action ``cutting paper'' is defined as follows: The initial action of ``cutting paper'' is to take scissors from the table and the final action is to put it on the table. We recruited 20 people to perform these tasks. 
All users were right handed. Our wrist-mounted camera and head-mounted camera dataset respectively produced 6.5 h (about 690,000 frames) of images.

\vspace{-2mm}
\subsection{Characteristics}
\begin{figure}[t]
\begin{center}
\includegraphics[width=0.87\linewidth]{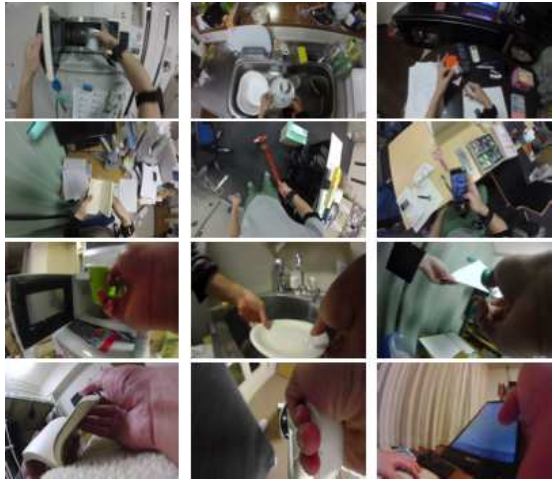}
\end{center}
\vspace{-4mm}
   \caption{Example images from our head-mounted dataset (top-half) and wrist-mounted dataset (bottom-half). We present a wide variety of scenes and ADL classes.}
\label{fig:da_sample}
\vspace{-4mm}
\end{figure}

\vspace{-2mm}
Various objects are handled in daily life. Therefore, for ADL recognition, it is important to be able to recognize them in diverse environments. For this study, we asked users to take videos in their own homes. As shown in the examples in Figure \ref{fig:da_sample}, the environments caught on camera differ depending on the user. More examples are shown in the Supplemental Materials.

\vspace{-3mm}
\section{Experiments}
\vspace{-2mm}
\subsection{Experiment protocols}
\vspace{-2mm}
\label{sec:protocols}
We used 16-layer VGG net \cite{Simonyan14c} pre-trained on the ImageNet 2012dataset \cite{deng2009imagenet} for the CNN architecture in the same manner as the LCD \cite{xu2015discriminative}. Motion features are also important in action recognition. Therefore, we evaluated our dataset not only with CNN descriptors but also with iDT \cite{wang2013action}. Following \cite{wang2013action}, we reduced the dimensions of the descriptors (HOG, HOF, and MBH) by a factor of 2 with PCA and encode them with FV, where the component number of the Gaussian mixture model was 256. We applied power and L2 normalization to aggregated vectors. As a classifier, we used a one-vs.-all SVM with linear kernel, setting $C=100$. We used leave-one-user-out cross-validation for the evaluation so that the same person does not appear across both training and test data for ADL recognition. The iteration number of our methods is fixed at five because it usually converges in a few iterations.

\vspace{-3mm}
\subsection{Evaluating DSTAR and our dataset}
\vspace{-2mm}

\begin{table}[tb]
 \begin{center}
 \small
  \begin{tabular}{lcc}
  \toprule
Video Features				&WCD		&HCD		\\ \hline
LCD+VLAD \cite{xu2015discriminative}			&78.6		&62.4		\\ 
LCD{\small spp}+VLAD \cite{xu2015discriminative}	&73.4		&51.3		\\ 
DSAR (ours)				&{\bf 82.0}		&61.6		\\ 
DSTAR (ours)				&{\bf 83.7}		&62.0		\\
 STAR$^{\ast}$				&77.0		&53.5 \\
  \bottomrule
  \end{tabular}
 \end{center}
  \vspace{-3mm}
  \caption{Mean classification accuracy of the proposed methods on the wrist-mounted camera dataset (WCD) and the head-mounted camera dataset (HCD). STAR$^{\ast}$ is the method without weight optimization, which is equivalent to a spatiotemporal pyramid \cite{laptev2008learning}.}
 \label{tab:stoal_and_camera}
  \vspace{-1mm}
\end{table}

We evaluated our approach on our wrist-mounted camera dataset (WCD) and head-mounted camera dataset (HCD). For fair comparison, we reduced the LCD dimensions from 512-D to a various range of dimensions such as 64-D, 128-D, and 256-D with PCA, and encoded them with various numbers of centers $K$ in VLAD such as $K=64, 128, 256, 512, 1024$ as in \cite{xu2015discriminative} to find the best ones. We also explored the best choice of dimensions, $K$, $N_{{\rm sp}}$, and $N_{{\rm tmp}}$, for our method. We describe the best parameters and how they are determined in the Supplemental Materials because of the limited space here.

Table \ref{tab:stoal_and_camera} presents the action classification accuracy of our dataset. Comparing the cameras, we found the accuracy on WCD to  be superior to that on HCD for every method. Next, we compared each method on WCD. Actually, DSAR, which retains spatial information after encoding, showed superior performance to LCD; DSTAR, which retains not only spatial information but also temporal information, exhibited superior performance to DSAR on both datasets. Results showed that an LCD with a spatial pooling layer (LCD{\small spp}) did not improve performance on our dataset, unlike TRECVID MEDTest 13 and 14 \cite{trecvid13, trecvid14}. The images captured by a wrist-mounted camera have strong spatial bias, and ADL actions have weak temporal bias, as described in Section \ref{sec:wmc}. From the obtained results, we can confirm that using spatial and temporal bias improves recognition accuracy on WCD. However, DSAR and DSTAR did not improve HCD performance. As shown in Figure \ref{fig:wrist_mean}, we cannot confirm strong spatial bias in the images captured by a head-mounted camera. Cutting features in every cell only made the features sparse. Aggregated video representation does not get more discriminative than without cutting if images have no strong spatial bias. Consequently, DSAR and DSTAR can be shown to improve recognition accuracy more for WCD than for HCD.  Through these recognition results, we can confirm that using a wrist-mounted camera and considering spatial and temporal information improved ADL recognition performance.

\begin{figure}[t]
\begin{center}
   \includegraphics[width=0.78\linewidth]{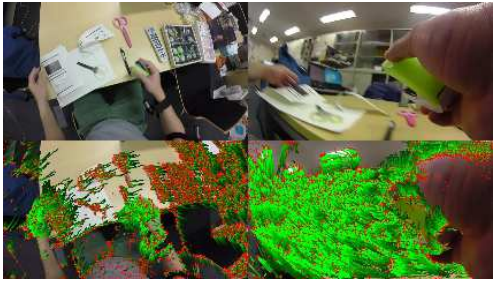}
\end{center}
\vspace{-2mm}
   \caption{Visualization example of iDT on a head-mounted camera ({\bf left}) and a wrist-mounted camera ({\bf right}). These images were captured simultaneously. Green lines are trajectories that were removed from the backgrounds with iDT. It is apparent that many background points in the image of a wrist-mounted camera are regarded as the foreground because of large motion on the camera.}
\vspace{-2mm}
\label{fig:visualized_idt}
\end{figure}

\vspace{-4mm}
\subsection{Applicability in existing datasets}
\vspace{-2mm}
Table \ref{tab:other} shows how wide our methods can be applied. 
We first evaluated LCD and our methods on UCIADL \cite{pirsiavash2012detecting}. As shown in the table, our methods did not improve the performance since this dataset has low spatial bias. Moreover, we evaluated them on UCF101 \cite{soomro2012ucf101}, which is one of the representative datasets of typical action recognition. Although not as strong as our wrist-mounted dataset, this dataset has substantial bias. Therefore, DSTAR showed better performance on this dataset than LCD did. Details are shown in the Supplemental Materials.

\begin{table}[tb]
 \begin{center}
 \small
  \begin{tabular}{lccc}
  \toprule
Dataset								&LCD \cite{xu2015discriminative}	&DSAR	&DSTAR	\\ \hline
UCIADL \cite{pirsiavash2012detecting}		&{\bf 73.7}		&71.8	&72.6	\\
UCF101 \cite{soomro2012ucf101}			&76.8		&78.7	&{\bf 79.3}	\\
MILADL (WCD)							&78.6		&82.0	&{\bf 83.7}	\\
  \bottomrule
  \end{tabular}
 \end{center}
  \vspace{-3mm}
  \caption{Mean classification accuracy on existing datasets.}
 \label{tab:other}
 \vspace{-2mm}
\end{table}

\vspace{-2mm}
\subsection{Fusing motion features and cameras}
\vspace{-2mm}
\begin{table}[tb]
 \begin{center}
 \small
  \begin{tabular}{lcc}
  \toprule
Video Features				&WCD			&HCD		\\ \hline
iDT+FV \cite{wang2013action}	&73.6			&78.1		\\ 
LCD \& iDT+FV				&84.1			&80.5		\\ 
DSTAR \& iDT+FV			&{\bf 85.5}			&80.2	\\ \hline
DSTAR (WCD) \& iDT+FV (HCD)&\multicolumn{2}{c}{{\bf 89.7}} \\
  \bottomrule
  \end{tabular}
 \end{center}
 \vspace{-3mm}
  \caption{Mean classification accuracy of combining CNN-based descriptors with motion features, and a wrist-mounted camera with a head-mounted camera.}
 \label{tab:fusion_adl}
  \vspace{-4mm}
\end{table}

From Tables \ref{tab:stoal_and_camera} and \ref{tab:fusion_adl}, we can confirm that iDT features are less discriminative than CNN-based features on WCD.

Comparing the iDT on both cameras, we found that iDT on HCD showed better performance than on WCD unlike the CNN descriptors. We can ascertain this reason from Figure \ref{fig:visualized_idt}. As the figure shows, iDT failed to remove the backgrounds from the video captured by a wrist-mounted camera compared with a head-mounted camera because of the large motion of the camera. Therefore, iDT on HCD is superior to that on WCD. 

Jain \etal \cite{jain201515} showed that combining object features extracted by CNN with motion features such as iDT boosts action classification accuracy. Following their conclusion, we also demonstrate how our method was affected by the combination of motion features. We fused our methods with iDT on each dataset by simply averaging the score obtained using our methods and the mean score obtained by all iDT scores. As Table \ref{tab:fusion_adl} shows, the performance of motion features was boosted by our methods more than by LCD. Although iDT features were more discriminative on HCD than on WCD, the combined features showed better performance on WCD than on HCD. Unlike action recognition, object features are more effective than motion features in ADL recognition because the critical key is the handled object. Therefore, we can find that wrist-mounted cameras are more suitable for ADL recognition than head-mounted cameras.

In case the user wears both a head-mounted camera and a wrist-mounted camera, we can choose superior information from wrist-mounted cameras and head-mounted cameras. Better object information is obtainable from wrist-mounted cameras, but better motion information is obtainable from head-mounted cameras. Therefore, we combined DSTAR on WCD with iDT on HCD to achieve the best accuracy of 89.7\% on our dataset.

\begin{figure}[t]
\begin{center}
   \includegraphics[width=0.75\linewidth]{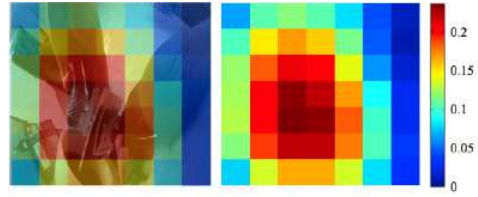}
\end{center}
\vspace{-4mm}
   \caption{Visualization of DSTAR {\it spatial} weights on the wrist-mounted camera dataset \textcolor{red}{\protect \footnotemark[4]}.}
\vspace{-4mm}
\label{fig:spatial_weights}
\end{figure}

\begin{figure}[t]
\begin{center}
   \includegraphics[width=0.75\linewidth]{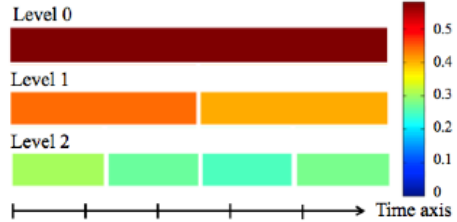}
\end{center}
\vspace{-4mm}
   \caption{Visualization of DSTAR {\it temporal} weights on the wrist-mounted camera dataset \textcolor{red}{\protect \footnotemark[4]}.}
\vspace{-4mm}
\label{fig:temporal_weights}
\end{figure}
\footnotetext[4]{These weights were obtained when we dealt with subject nos. 2--20 for training and no. 1 for testing.}

\vspace{-3mm}
\section{Discussion}
\vspace{-2mm}
\subsection{Visualized weights of DSTAR}
\vspace{-2mm}
Figures \ref{fig:spatial_weights} and \ref{fig:temporal_weights} show the absolute values of spatial and temporal weights $W_{{\rm sp}}$ and $W_{{\rm tmp}}$ calculated using DSTAR on WCD. This figure presents the optimal discriminative weights for the respective cells.

{\bf Spatial weight:} 
In Figure \ref{fig:spatial_weights}, it is apparent that cells near the center are important for classification, whereas cells on the right side are less important. The user's palm always appears. No object appears on the right side of a wrist-mounted camera image. Therefore, the right side area in the image has less information for recognition. The features obtained from the upper left cell and the bottom left cell are also less discriminative because backgrounds unrelated to the user's action are often captured in these cells. However, handled objects often appear in the middle area of a wrist-mounted camera image. Discriminative features can be obtained from the cells of these areas.

{\bf Temporal weight:} Although not as strong as the spatial bias of the image captured by a wrist-mounted camera, each ADL class has weak temporal bias. As Figure \ref{fig:temporal_weights} shows, although full-length features (level 0) are the most important, temporally cut features (levels 1 and 2) have different weights. Using temporally cut pyramids improved the recognition performance, as presented in Table \ref{tab:stoal_and_camera}. Additionally, slight differences are apparent at the same level. At level 2, the beginning and the end of the action are slightly more important than the middle of the action.

\begin{figure*}[t]
\begin{center}
   \includegraphics[width=0.75\linewidth]{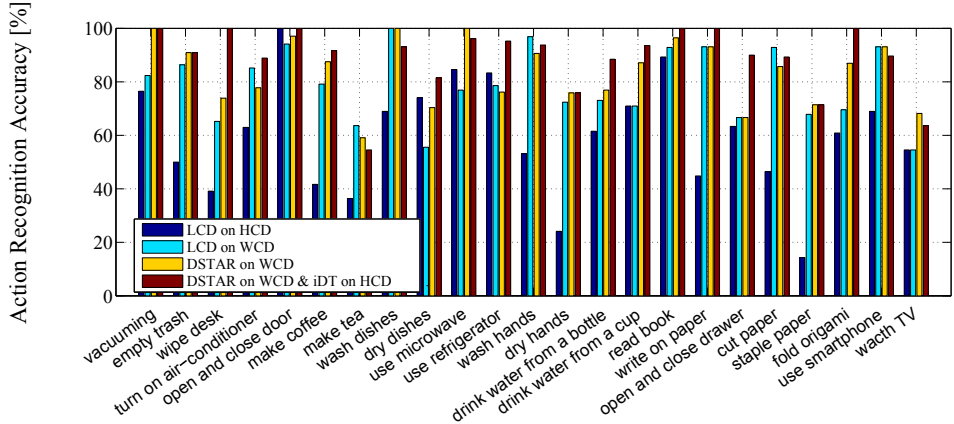}
\end{center}
\vspace{-5mm}
   \caption{This figure represents the recognition accuracy of each ADL class. It also shows the differences between the models.}
   \label{fig:class_result}
\vspace{-3mm}
\end{figure*}

\vspace{-3mm}
\subsection{Analysis of ADL classification results}
\vspace{-2mm}
We analyze the classification results here. Figure \ref{fig:class_result} presents the results of four different methods: LCD on HCD, LCD on WCD, DSTAR on WCD, and, finally, DSTAR on WCD and iDT on HCD. 

{\bf Comparing HCD with WCD:} We first compared both cameras using LCD. Results showed that 18 classes showed superior performance on WCD over HCD; these classes improved by 28.1\% on average. Especially, ``write on paper,'' ``cut paper,'' and ``staple paper'' were improved significantly. These classes are actions wherein users use small objects such as a pen, scissors, and a stapler. A head-mounted camera captures these objects at a small scale. However, a wrist-mounted camera can capture large-scale images even of small objects. Four classes, however, showed inferior performance on WCD compared to HCD, and ``dry dishes'' is the class in which the classification accuracy declined considerably: 18.5\%. On WCD, ``dry dishes'' was more often confused with ``wash dishes'' than on HCD, although all actions of ``wash dishes'' were recognized correctly.

{\bf Comparing LCD with DSTAR:} Next, we compared DSTAR with LCD on WCD. Using DSTAR instead of LCD improved the accuracy on 14 classes. Its average improvement rate was 10.1\%. One significantly improved action class was ``vacuuming.'' When we used a vacuum cleaner with a wrist-mounted camera, the floor and other unrelated backgrounds appeared on the left side of the wrist-mounted camera image. Actually, DSTAR was considered to improve the performance by reducing the importance of the features extracted from these areas. On five classes, the accuracy decreased, but the average rate of decrease was only 5.5\%.

{\bf Adding iDT on HCD with DSTAR on WCD: } Finally, we noticed how adding iDT affected HCD. Actually, 14 classes improved their performance by adding iDT on HCD; these classes improved by 10.4\% on average. Especially, ``wipe desk'' was highly improved. Because the object information of ``wipe desk'' on HCD was often confused with other actions done near the table, using a wrist-mounted camera improved the performance. However, the ``wipe desk'' motion was distinctive. Consequently, adding iDT on HCD boosted the performance again. Only on five classes did the accuracy decline, the average of which was only 4.7\%. In case the user wears both cameras, we can obtain better recognition.

\vspace{-3mm}
\section{Conclusion}
\vspace{-2mm}
This study examined the recognition of ADL with a wrist-mounted camera. We developed a publishable dataset of videos taken with a head-mounted camera and a wrist-mounted camera. Additionally, we proposed a novel video representation that aggregated CNN descriptors spatially and temporally, and optimized their weights both iteratively and alternately. Finally, using the proposed dataset, we quantitatively demonstrated the benefits of a wrist-mounted camera over a head-mounted camera and those of our proposed method over previous methods. We believe that our work will help spread the use of cameras attached to wrist-mounted devices.

 \cleardoublepage
 {\small
\bibliographystyle{ieee}
\bibliography{egpaper_for_review}
}

\onecolumn

\section*{Supplemental Material}

\setcounter{section}{0}
\renewcommand{\thesection}{\Alph{section}}
\section{Calculating PLS to compute eigenvectors that construct a weight matrix $W_{\rm sp}$}

We compute eigenvectors that construct $W_{\rm sp}$ by calculating partial least squares (PLS). 
The idea to use PLS for obtaining weights for discriminative features was derived from \cite{harada2011discriminative}.
Here, we reproduce Eq. (3), (4), and (5) in our paper below:
\begin{eqnarray}
  V &=& \sum_{i=1}^a \sum_{j=1}^a \vector{v}(i,j) \vector{w}_{( i, j )}^\top = V_{\rm sp} W_{\rm sp}, \\
  V_{\rm sp} &=& ( \vector{v}(1,1), \vector{v}(1,2), \ldots, \vector{v}(a,a)), \\
  W_{\rm sp} &=& ( \vector{w}_{(1,1)}, \vector{w}_{(1,2)}, \ldots, \vector{w}_{(a,a)})^\top.
  \label{eq:weights}
\end{eqnarray}
We let $\vector{w} \in \mathbb{R}^{a^2}$ be a column vector in $W_{\rm sp}$ and $\vector{x} \in \mathbb{R}^{MK}$ denote the corresponding column vector in $V$ (\ie $\vector{x} = V_{\rm sp} \vector{w}$).
Suppose that we have $N$ labeled training samples $\{\vector{x}_i,y_i\}_{i=1}^N$ with $C$ classes, where $\vector{x}_i = V_{\rm sp}^i \vector{w}$ and $y_i$ represents the class label of the $i$-th training sample ranging from 1 to $C$.
The between-class covariance matrix $S_b$ can be written as follows:
\begin{equation}
  S_b = \frac{1}{N} \sum_{c=1}^C n_c ( \bar{\vector{x}}_c - \bar{\vector{x}} ) ( \bar{\vector{x}}_c - \bar{\vector{x}} )^\top,
\end{equation}
where $\bar{\vector{x}}_c = \frac{1}{n_c} \sum_{i \in \{ i | y_i=c\}} \vector{x}_i$, $\bar{\vector{x}} = \frac{1}{N} \sum_i \vector{x}_i$, and $n_c$ is the number of samples in the $c$-th class.
The trace of $S_b$ is given by:
\begin{equation}
  {\rm tr}S_b = \vector{w}^\top \Sigma_b \vector{w}, \label{eq:trace}
\end{equation}
where
\begin{equation}
  \Sigma_b = \frac{1}{N} \sum_{c=1}^C n_c (M_c - M)^\top (M_c - M).
\end{equation}
Here, $M_c = \frac{1}{n_c} \sum_{i \in \{ i | y_i=c\}} \vector{x}_i$ is the mean of $\vector{x}_i$ belonging to the $c$-th class, 
and $M = \frac{1}{N} \sum_i \vector{x}_i$ is the mean of all samples in the training dataset.
By maximizing Eq. (\ref{eq:trace}) under the condition $\vector{w}^\top \vector{w} = 1$, we obtain the eigenvector of the following eigenvalue problem:
\begin{equation}
  \Sigma_b \vector{w} = \lambda \vector{w},
\end{equation}
where $\lambda$ is the eigenvalue corresponding to the eigenvector $\vector{w}$.
We select the $N_{\rm sp}$ largest eigenvalues $\lambda_1, \dots, \lambda_{N_{\rm sp}}$, and the corresponding eigenvectors $\vector{w}_1, \dots, \vector{w}_{N_{\rm sp}}$.
Finally, we create $W_{\rm sp}$ by arranging $\vector{w}_1, \dots, \vector{w}_{N_{\rm sp}}$ in a row.
As described in our paper, $W_{\rm tmp}$ can be obtained in the same manner as $W_{\rm sp}$ when $W_{\rm sp}$ is fixed.

\clearpage
\section{Action definition of our dataset }
Table \ref{tab:definition} shows the definition of each class in our dataset.

\begin{table}[htpb]
\begin{center}
\small
\begin{tabular}{|l||p{13cm}|}
\hline
Activity class & Definition\\
\hline\hline
1. Vacuuming				&The initial action is taking a vacuum cleaner and the final action is putting it back after vacuuming. \\
2. Empty trash 				&The initial action is taking a garbage bag and the final action is tying it. \\
3. Wipe desk				&The initial action is taking a kitchen cloth, the next action is wiping, and the final action is putting it after washing.\\
4. Turn on air-conditioner		&The initial action is taking a remote controller, the next action is turing on, and the final action is making sure an air-conditioner turns on.\\
5. Open and close door		&The initial action is taking door knob and the final action is releasing user's hand.\\
6. Make coffee				&The initial action is taking a drip filter and the final action is throwing it away after pouring.\\	
7. Make tea				&The initial action is taking tea bag and the final action is throwing it away after pouring.\\
8. Wash dishes				&The initial action is turning a tap on and the final action is turning it off after washing. \\
9. Dry dishes				&The initial action is taking a cloth and the final action is putting it after drying.\\
10. Use microwave			&The initial action is putting a thing into microwave and the final action is opening microwave and taking the thing out.\\
11. Use refrigerator			&The initial action is opening a refrigerator door, the next action is taking a thing, and the final action is closing the door.\\
12. Wash hands			&The initial action is turning a tap on and the final action is turning it off.\\
13. Dry hands				&The initial action is taking a cloth and the final action is putting the cloth after drying.\\
14. Drink water from a bottle	&The initial action is taking a plastic bottle and the final action is putting it after drinking.\\
15. Drink water from a cup 	&The initial action is taking a cup and the final action is putting it after drinking.\\
16. Read book				&The initial action is taking a book and the final action is putting it after reading.\\
17. Write on paper			&The initial action is taking a pen and the final action is putting it after writing.\\
18. Open and close drawer	&The initial action is opening a drawer, the next action is taking a thing, and the final action is closing the drawer.\\
19. Cut paper				&The initial action is taking scissors and the final action is putting them after cutting. \\
20. Staple paper 			&The initial action is taking stapler and the final action is putting it after stapling. \\
21. Fold origami 			&The initial action is taking origami and the final action is folding it.\\
22. Use smartphone 			&The initial action is taking a smartphone and the final action is putting it after using.\\
23. Watch TV				&The initial action is taking a remote controller and the final action is turning TV off after watching.\\
\hline
\end{tabular} 
\end{center}
\caption{Activity definitions of our dataset}
\label{tab:definition}
\end{table}

\clearpage
\section{Confusion matrix}
\begin{figure}[h]
\begin{center}
\begin{tabular}{c}
\begin{minipage}{0.5\hsize}
\begin{center}
   \includegraphics[clip, width=\linewidth]{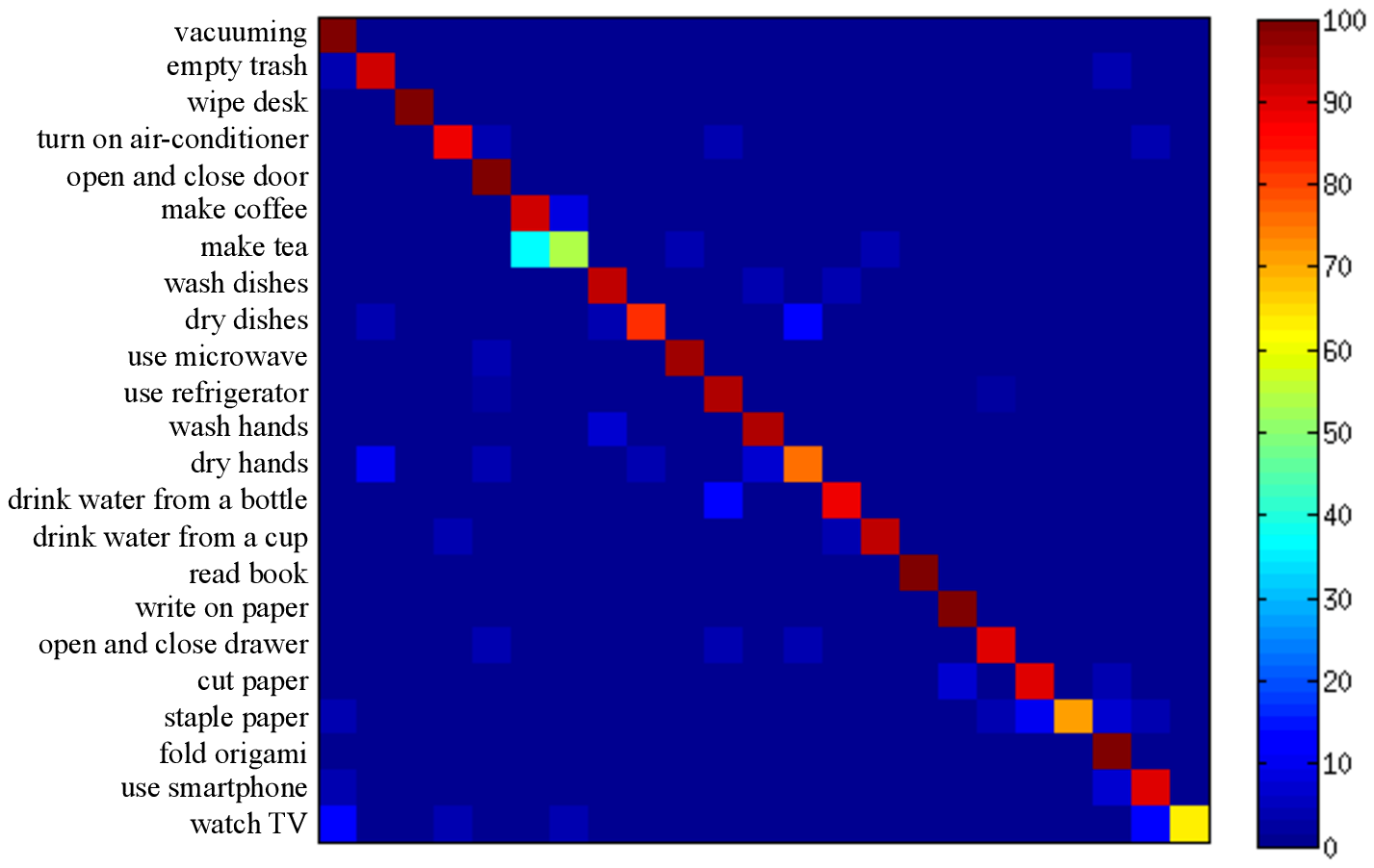}
\end{center}
\vspace{-10mm}
\caption{The confusion matrix for LCD on HCD}
\label{fig:conf1}
\end{minipage}

\begin{minipage}{0.5\hsize}
\begin{center}
   \includegraphics[clip, width=\linewidth]{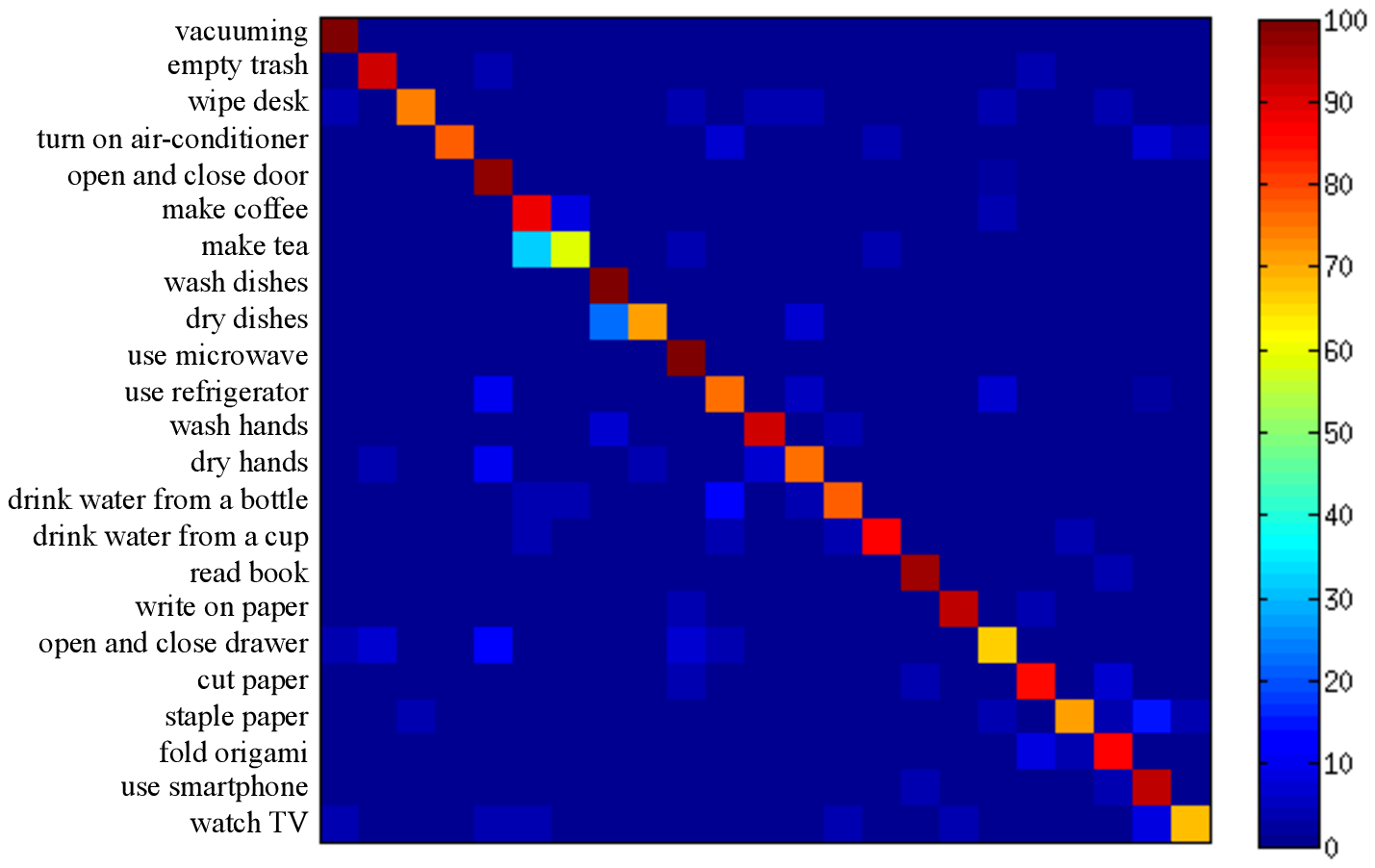}
\end{center}
\vspace{-10mm}
\caption{The confusion matrix for LCD on WCD}
\label{fig:conf2}
\end{minipage}
\end{tabular}
\end{center}
\end{figure}

\begin{figure}[h]
\begin{center}
\begin{tabular}{c}

\begin{minipage}{0.5\hsize}
\begin{center}
   \includegraphics[clip, width=\linewidth]{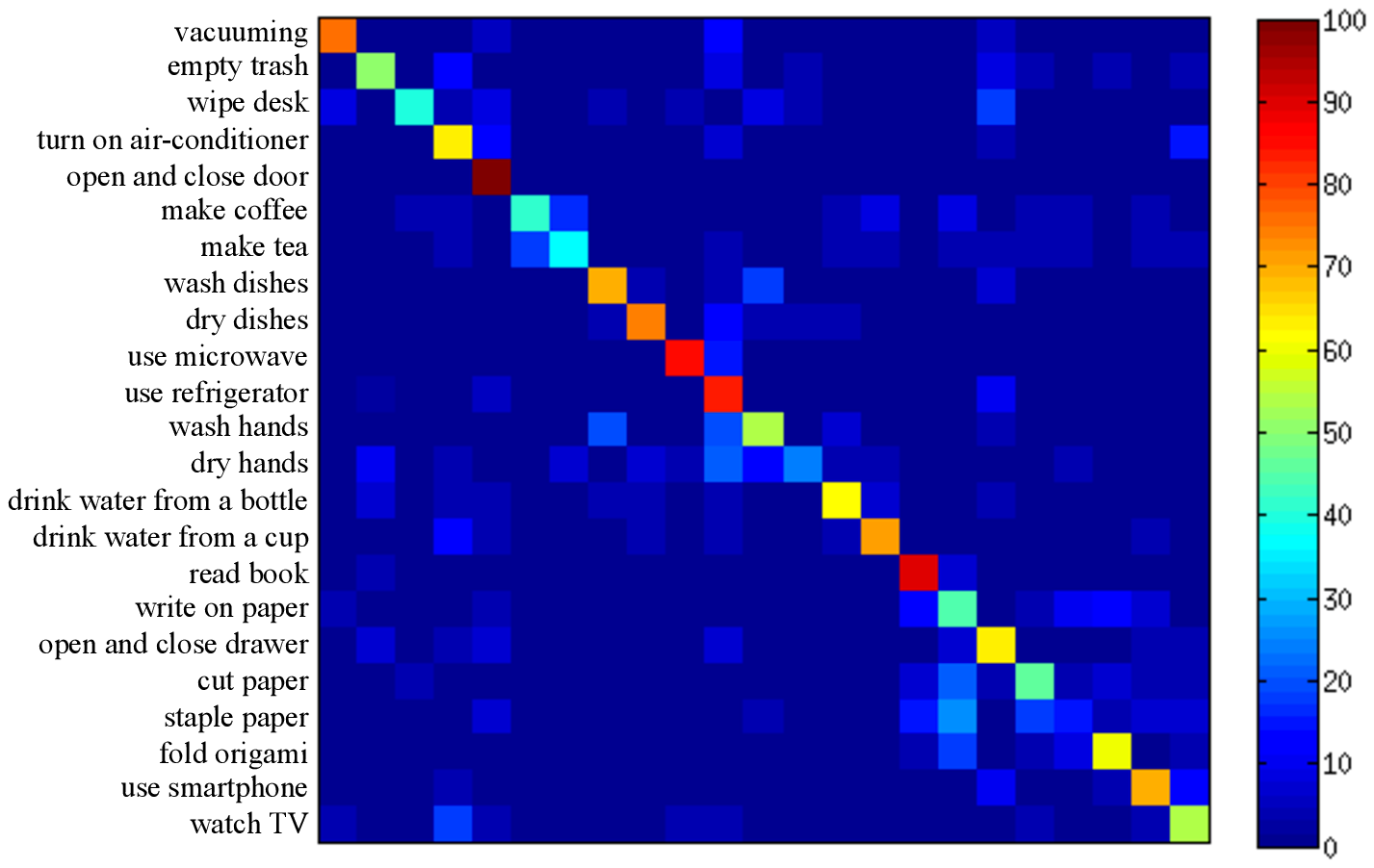}
\end{center}
\vspace{-10mm}
\caption{The confusion matrix for DSTAR on WCD}
\label{fig:conf3}
\end{minipage}

\begin{minipage}{0.5\hsize}
\begin{center}
   \includegraphics[clip, width=\linewidth]{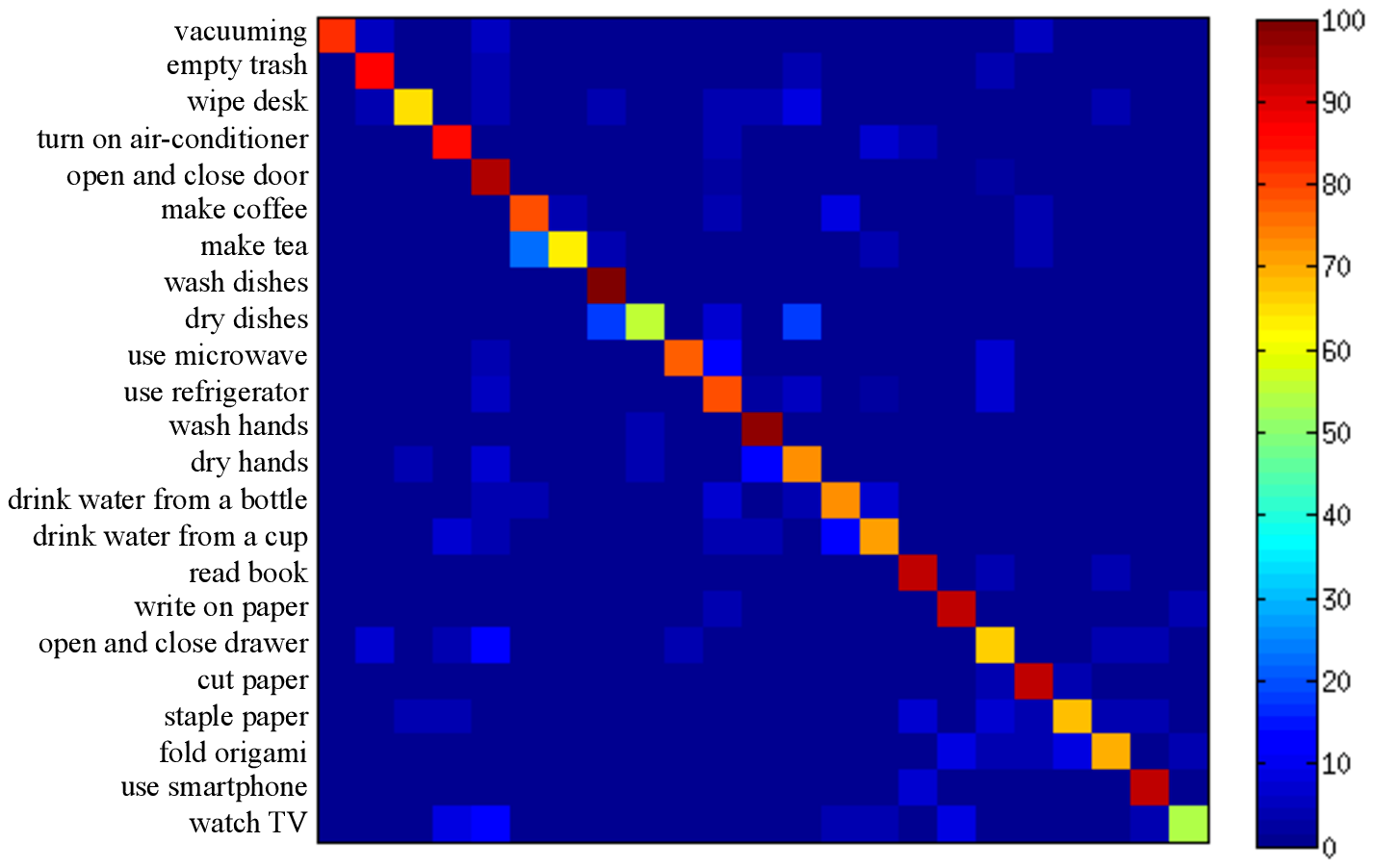}
\end{center}
\vspace{-10mm}
\caption{The confusion matrix for DSTAR on WCD \& iDT on HCD}
\label{fig:conf4}
\end{minipage}

\end{tabular}
\end{center}
\end{figure}

Figures \ref{fig:conf1}, \ref{fig:conf2}, \ref{fig:conf3}, and \ref{fig:conf4} are the confusion matrices.
We can see how action classes are confused in each figure.
For example, ``make coffee'' and ``make tea'' are confused in every case. These actions have common handled objects such as a mug and pot. The biggest difference is whether the user uses tea bag or coffee beans and filter. It is difficult for head-mounted camera to recognize these objects. However, wrist-mounted camera can recognize small handed objects easily. Thus, LCD \cite{xu2015discriminative}  on wrist-mounted camera dataset (WCD) recognizes ``make coffee'' and ``make tea'' better than LCD on head-mounted camera dataset (HCD).


\clearpage
\section{Analysis of the impact of parameters}
In this section, we find the best parameters for each method.

\subsection{Parameters for LCD}
We first find the best parameters for LCD; the number of centers $K$ in VLAD and descriptor dimension. 
Table \ref{tab:LCD on WCD} shows that features get more discriminative with the increase of $K$ on WCD. However when $K=1024$, the features get too sparse and less discriminative. Though we find the best parameter $(K, D)=(128, 256)$, the compressed dimension by PCA dose not seem to have much effect.

Table \ref{tab:LCD on HCD} also shows that features get more discriminative with the increase of $K$ on HCD. However when $K=512$, the features get too sparse and less discriminative. Unlike WCD, when dimensions of each descriptor are compressed from 512-D to 64-D, they lose the discriminative ability. This is understood as follows: the images captured by wrist-mounted camera have less variety than the images by head-mounted camera. Thus, the descriptors extracted from WCD can be more compact than those from HCD.

\begin{table}[h]
\begin{center}
\small
  \begin{tabular}{|c|c|c|c|c|c|}
\hline
 \backslashbox{Dimensions}{Clusters}		&$K=64$	&$K=128$	&$K=256$	&$K=512$	&$K=1024$ 	\\ \hline
64-D					&74.0		&74.1		&76.0		&78.2		&77.1	\\ \hline
128-D				&73.4		&74.3		&75.8		&{\bf 78.6}		&77.0	\\  \hline
256-D				&74.5		&76.8		&77.1		&78.5		&77.2	\\  \hline
  \end{tabular}
   \end{center}
     \caption{Impact on dimensions and numbers of centers $K$ for LCD on WCD}
     \label{tab:LCD on WCD}
\end{table}

\begin{table}[h]
\begin{center}
\small
  \begin{tabular}{|c|c|c|c|c|c|}
\hline
 \backslashbox{Dimensions}{Clusters}		&$K=64$	&$K=128$	&$K=256$	&$K=512$	&$K=1024$ 	\\ \hline
64-D				&48.5	&52.5	&57.6	&56.4	&57.0	\\ \hline
128-D			&54.9	&56.0	&{\bf 62.4}	&62.1	&61.2	\\ \hline
256-D			&56.1	&60.8	&60.1	&61.0	&61.5	\\ \hline
  \end{tabular}
   \end{center}
   \caption{Impact on dimensions and numbers of centers $K$ for LCD on HCD}
   \label{tab:LCD on HCD}
\end{table}

We also find the best parameter for LCD$_{{\rm SPP}}$. Tables \ref{tab:LCD SPP on WCD} and \ref{tab:LCD SPP on HCD} show the obtained results of LCD$_{{\rm SPP}}$. We can see similar trend as LCD without Spatial Pooling Pyramid (SPP) layer shown in Tables  \ref{tab:LCD on WCD} and  \ref{tab:LCD on HCD}. The best parameter $(K, D)$ are $(128, 256)$ for LCD$_{{\rm SPP}}$ on WCD and $(256,256)$ for {LCD$_{{\rm SPP}}$ on HCD. We employ the score obtained with these parameters in submitted paper.

\begin{table}[h]
\begin{center}
\small
  \begin{tabular}{|c|c|c|c|c|}
\hline
 \backslashbox{Dimensions}{Clusters}	&$K=64$		&$K=128$	&$K=256$	&$K=512$	\\ \hline
64-D				&65.9	&68.6	&70.9	&72.0	\\ \hline
128-D			&70.0	&70.2	&71.9	&73.2	\\ \hline
256-D			&73.3	&73.1	&{\bf 73.4}	&72.9	\\ \hline
  \end{tabular}
   \end{center}
     \caption{Impact on dimensions and numbers of centers $K$ for LCD$_{{\rm SPP}}$ on WCD}
     \label{tab:LCD SPP on WCD}
\end{table}

\begin{table}[h]
\begin{center}
\small
  \begin{tabular}{|c|c|c|c|c|}
\hline
 \backslashbox{Dimensions}{Clusters}	&$K=64$		&$K=128$	&$K=256$	&$K=512$	\\ \hline
64-D				&45.9	&46.3	&45.9	&45.9	\\ \hline
128-D			&46.6	&47.0	&48.7	&46.1	\\ \hline
256-D			&48.5	&46.9 	&{\bf 51.3}	&48.1	\\ \hline
  \end{tabular}
   \end{center}
\caption{Impact on dimensions and numbers of centers $K$ for LCD$_{{\rm SPP}}$ on HCD}
     \label{tab:LCD SPP on HCD}
\end{table}

\subsection{Number of spatial elements}
\label{sec:DSAR}
Next, we find the best parameters for DSAR; the number of centers $K$ in VLAD, descriptor dimension, and $N_{{\rm sp}}$. Tables \ref{tab:DSAR on WCD} and \ref{tab:DSAR on HCD} show the best parameter for DSAR on WCD and DSAR on HCD. We can see that $N_{{\rm sp}}$ dose not need to be a large number though it can be set up-to 49 in VGG-net \cite{Simonyan14c}  case. The similar trend can be seen in the D-SPR \cite{harada2011discriminative}. If features are cast into well-isolated space by PLS, using too large $N_{{\rm sp}}$ means adding inefficient features. 

For numbers of clusters, $K=512$ seems too sparse unlike Table \ref{tab:LCD on WCD}. We calculate weights $W_{{\rm sp}}$, shown in Eq. (\ref{eq:weights}), from separately aggregated features in each cell. These separately aggregated features can be more sparse than LCD features. Thus, the best number of clusters for DSAR is smaller than that of LCD. 

We can find the best parameter $(K, D, N_{{\rm sp}})=(64, 256, 5)$ for DSAR on WCD and $(K, D, N_{{\rm sp}})=(256, 128, 5)$ for DSAR on HCD from Tables \ref{tab:DSAR on WCD} and \ref{tab:DSAR on HCD}.

\begin{table}[h]
\begin{center}
\small
  \begin{tabular}{|c|ccc|ccc|ccc|ccc|}
\hline
Clusters	& \multicolumn{3}{c|}{$K=64$}	& \multicolumn{3}{c|}{$K=128$}	& \multicolumn{3}{c|}{$K=256$}	& \multicolumn{3}{c|}{$K=512$}		\\ \hline
\backslashbox{Dimensions}{$N_{{\rm sp}}$}		&5		&10		&20		&5		&10		&20		&5		&10		&20		&5		&10		&20			\\ \hline
64-D			&77.9	&78.8	&76.9	&78.8	&78.2	&77.4	&80.2	&78.7	&75.8	&77.4	&78.6	&74.2		\\ \hline
128-D		&80.2	&79.3	&77.9	&80.4	&81.4	&78.9	&81.1	&81.1	&77.1	&80.9	&79.9	&76.1		\\ \hline
256-D		&{\bf 82.0}	&81.0	&80.0	&81.6	&81.0	&79.3	&79.7	&80.5	&78.1	&80.6	&78.6	&76.7		\\ \hline
  \end{tabular}
   \end{center}
   \caption{Impact on dimensions, numbers of centers $K$, and $N_{{\rm sp}}$ for DSAR on WCD.}
   \label{tab:DSAR on WCD}
\end{table}

\begin{table}[h]
\begin{center}
\small
  \begin{tabular}{|c|ccc|ccc|ccc|ccc|}
\hline
Clusters	& \multicolumn{3}{c|}{$K=64$}	& \multicolumn{3}{c|}{$K=128$}	& \multicolumn{3}{c|}{$K=256$}	& \multicolumn{3}{c|}{$K=512$}		\\ \hline
\backslashbox{Dimensions}{$N_{{\rm sp}}$}		&5		&10		&20		&5		&10		&20		&5		&10		&20		&5		&10		&20			\\ \hline64-D			&58.6	&59.3	&57.2	&57.3	&55.6	&52.4	&57.1	&56.9	&53.4	&58.8	&56.9	&54.0		\\ \hline
128-D		&60.1	&57.3	&56.0	&60.2	&59.3	&56.4	&{\bf 61.6}	&58.4	&57.5	&60.9	&58.4	&56.7		\\ \hline
256-D		&59.7	&59.1	&57.8	&59.9	&59.3	&56.9	&61.5	&59.2	&56.3	&60.8	&59.2	&57.4		\\ \hline
  \end{tabular}
   \end{center}
  \caption{Impact on dimensions, numbers of centers $K$, and $N_{{\rm sp}}$ for DSAR on HCD.}
     \label{tab:DSAR on HCD}
\end{table}

\subsection{Number of spatial and temporal elements}
We finally find the best parameter for DSTAR; the number of centers $K$ in VLAD, descriptor dimension, and $N_{{\rm tmp}}$. Following the result described in Section \ref{sec:DSAR}, we fix $N_{{\rm sp}}=5$. Tables \ref{tab:DSTAR on WCD} and \ref{tab:DSTAR on HCD} show the best parameter for DSTAR on WCD and DSTAR on HCD. We can find the best parameter $(K, D, N_{{\rm tmp}})=(128, 64, 5)$ for DSAR on WCD and $(K, D, N_{{\rm sp}})=(128, 128, 5)$ for DSAR on HCD from Tables \ref{tab:DSAR on WCD} and \ref{tab:DSAR on HCD}.

\begin{table}[h]
\begin{center}
\small
  \begin{tabular}{|c|cc|cc|}
\hline
Clusters	& \multicolumn{2}{c|}{$K=64$}	& \multicolumn{2}{c|}{$K=128$}					\\ \hline
\backslashbox{Dimensions}{$N_{{\rm tmp}}$}	&3	&5	&3	&5				\\ \hline
64-D			&81.3&81.3				&82.8	&{\bf 83.7}		\\ 
128-D		&82.8&83.2				&82.6	&83.5		\\ \hline 
  \end{tabular}
   \end{center}
     \caption{Impact on dimensions, numbers of centers $K$, and $N_{{\rm tmp}}$ for DSTAR on WCD, with fixed $N_{{\rm sp}}=5$.}
     \label{tab:DSTAR on WCD}
\end{table}

\begin{table}[h]
\begin{center}
\small
  \begin{tabular}{|c|cc|cc|}
\hline
Clusters	& \multicolumn{2}{c|}{$K=64$}	& \multicolumn{2}{c|}{$K=128$}					\\ \hline
\backslashbox{Dimensions}{$N_{{\rm tmp}}$}	&3	&5					&3	&5				\\ \hline
64-D			&60.4&60.0				&58.7&59.4			\\ \hline
128-D		&60.2&59.6				&60.7&{\bf 62.0}		\\ \hline
  \end{tabular}
   \end{center}
     \caption{Impact on dimensions, numbers of centers $K$, and of $N_{{\rm tmp}}$ for DSTAR on HCD, with fixed $N_{{\rm sp}}=5$.}
     \label{tab:DSTAR on HCD}
\end{table}


\end{document}